\documentclass[letterpaper, 10 pt, conference]{ieeeconf}  

\IEEEoverridecommandlockouts                              

\usepackage{graphics} 
\usepackage{epsfig} 
\usepackage{mathptmx} 
\usepackage{times} 
\usepackage{amsmath} 
\usepackage{amssymb}  
\usepackage{comment}
\usepackage[T1]{fontenc}
\usepackage{multirow}
\usepackage{textcomp}
\usepackage{gensymb}
\usepackage{balance}
\usepackage{soul}
\usepackage{color}

\usepackage{adjustbox}
\usepackage[utf8]{inputenc}
\usepackage{csquotes}
\usepackage[english]{babel}

\usepackage[backend=biber,style=numeric,sorting=none, maxnames=99, giveninits=true]{biblatex} 
\usepackage[font=small,skip=5pt]{caption}
\title{\LARGE \bf
Geomorphological Analysis Using Unpiloted Aircraft Systems, \\ Structure from Motion, and Deep Learning
}

\author{Zhiang Chen, Tyler R. Scott, Sarah Bearman, Harish Anand, \\ Devin Keating, Chelsea Scott, J Ram\'{o}n Arrowsmith, Jnaneshwar Das
\thanks{Authors are affiliated with the School of Earth and Space Exploration, Arizona State University, 781 Terrace Mall, Tempe, AZ 85287, USA \newline
{\tt\small \{zch,ramon.arrowsmith,jnaneshwar.das\}@asu.edu}
        }%
}

\begin{document}

\maketitle
\thispagestyle{empty}
\pagestyle{empty}
\maxdeadcycles=200

\begin{abstract}

We present a pipeline for geomorphological analysis that uses structure from motion (SfM) and deep learning on close-range aerial imagery to estimate spatial distributions of rock traits (size, roundness, and orientation) along a tectonic fault scarp. The properties of the rocks on the fault scarp derive from the combination of initial volcanic fracturing and subsequent tectonic and geomorphic fracturing, and our pipeline allows scientists to leverage UAS-based imagery to gain a better understanding of such surface processes. We start by using SfM on aerial imagery to produce georeferenced orthomosaics and digital elevation models (DEM). A human expert then annotates rocks on a set of image tiles sampled from the orthomosaics, and these annotations are used to train a deep neural network to detect and segment individual rocks in the entire site. The extracted semantic information (rock masks) on large volumes of unlabeled, high-resolution SfM products allows subsequent structural analysis and shape descriptors to estimate rock size, roundness, and orientation. We present results of two experiments conducted along a fault scarp in the Volcanic Tablelands near Bishop, California. We conducted the first, proof-of-concept experiment with a DJI Phantom 4 Pro equipped with an RGB camera and inspected if elevation information assisted instance segmentation from RGB channels. Rock-trait histograms along and across the fault scarp were obtained with the neural network inference. In the second experiment, we deployed a hexrotor and a multispectral camera to produce a DEM and five spectral orthomosaics in red, green, blue, red edge, and near infrared. We focused on examining the effectiveness of different combinations of input channels in instance segmentation.



\end{abstract}

\section{INTRODUCTION}

Geographic Information Systems (GIS) have helped integrate a wide range of data sources, enabling efficient approaches for geological studies \cite{goodchild2010twenty}. Traditionally, field surveys have been a gold standard for data collection due to low bias and high tolerance for ambiguity. However, there are logistical constraints to field surveys, and findings may not be as unbiased as previously assumed \cite{salisbury2015validation}. Meanwhile, remote sensing for the collection of close-range terrestrial data has evolved from traditional methods such as airplanes, balloons, and kites equipped with cameras and LiDARs to the use of versatile robotic platforms such as Unpiloted Aerial Vehicles (UAV) or Unpiloted Aerial Systems (UAS)~\cite{das2013environmental, das2015devices}. Combining data collection with UAV/UAS and Structure from Motion (UAS-SfM) offers a low-cost solution for rapid mapping of geologic sites, and generates data products like digital surface models (DSM), digital elevation models (DEM), and cm-scale orthomosaics. Such data products require semi-automatic methods to yield interpretable information. 

In recent years, deep neural networks have demonstrated unprecedented success in image classification, segmentation, and object detection, leading to extensive application in satellite and airborne image analysis. Compared with deep learning applications with satellite imagery~\cite{zhu2017deep, kussul2017deep}, close-range UAS imagery with high resolution extends the use of deep learning to features of interest over a large range of feature sizes, ranging as low as a few centimeters. However, directly using camera perspective images from UAS does not provide precise georeference for features of interest, which is essential in some applications, such as geological studies.

\begin{figure}
\centering
\vspace{6pt}
\includegraphics[width=0.48\textwidth]{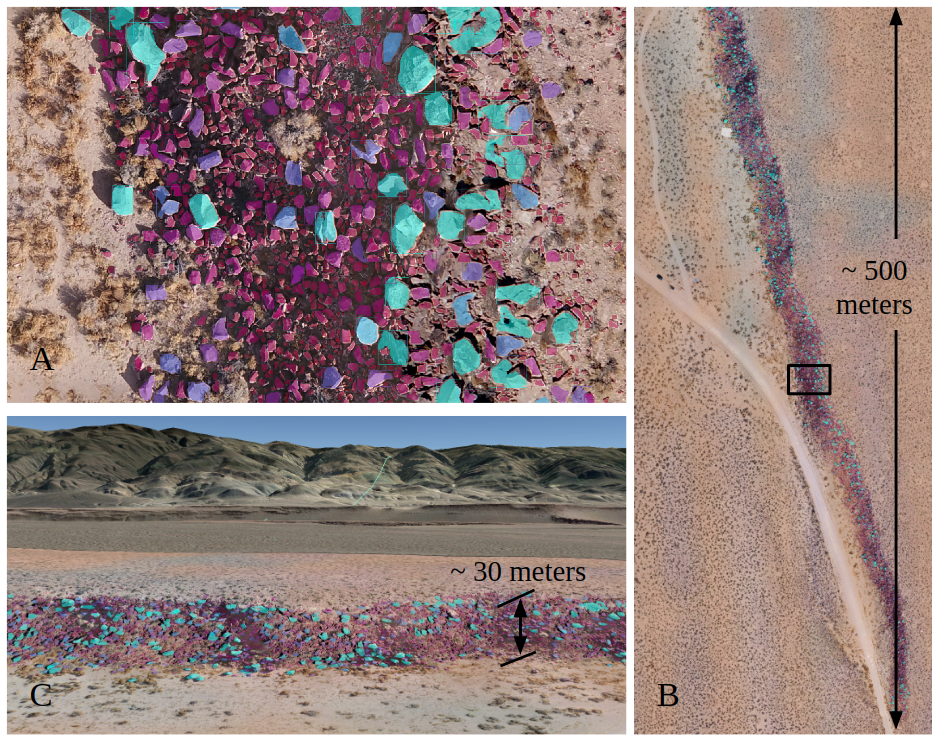}
\caption{Visualization of instance segmentation. Colors indicate rock sizes. (A) Partial enlargement of the black rectangle in B. (B) Instance segmentation at the study area I. (C) Ground view after loading the GeoTiff image into Google Earth. 
}
\label{fig:color-segmentation}
\vspace{-2mm}
\end{figure}

\begin{figure*} [tp]
\centering
\vspace{6pt}
\includegraphics[width=0.9\textwidth]{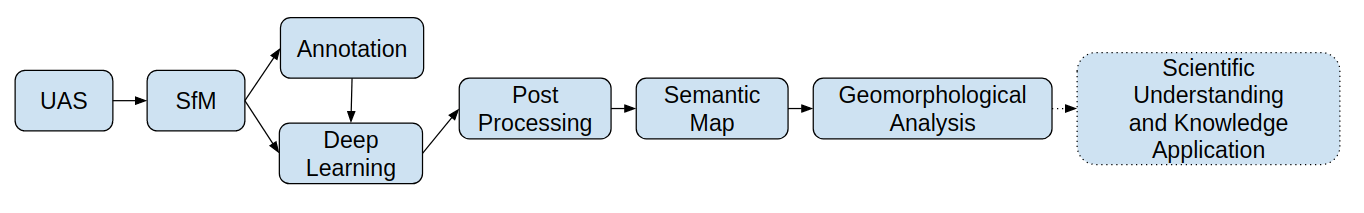}
\caption{Workflow of UAS-SfM-DL. The presented pipeline expands the utilities of models from UAS-SfM. }
\label{fig:pipeline}
\end{figure*}

Our work is motivated by the need for precise, large spatial-scale estimation of geomorphological features. In this study, we collect and process data that potentially correlate with surface processes of tectonic fault scarps in Volcanic Tablelands near Bishop, California \cite{ferrill2016observations}. Rock traits such as size, roundness, and orientation are of importance in many surface process studies, including earthquake geology research. Rock size distributions in the field site reflect both the initial cooling joint fracture geometry and the faulting-induced fracturing, which both vary with position as a function of strain magnitude and linkage characteristics. In addition, impacts during transport and thermal cycling may further drive fracturing and influence the particle sizes and shapes \cite{mcfadden2005physical}. Rock orientations indicate the character of downslope transport along the fault scarps, enhancing our understanding of erosional processes. Current analyses of topographic or imagery-based models produced by SfM largely rely on experts manually annotating features of interest (rocks).

We present a pipeline that combines UAS, SfM, and deep learning (UAS-SfM-DL) to produce high-resolution semantic maps of objects, such as rocks, with wide variation in size and appearance (Fig.~\ref{fig:pipeline}). We show that deep learning can be effectively applied to products from SfM, which may contain artifacts resulting from reconstruction. Advantages of the presented system include low-cost, rapid deployment and analysis, and automated processing with limited expert intervention. In comparison with deep learning methods on perspective UAS imagery~\cite{ammour2017deep}, the UAS-SfM-DL paradigm can produce consistent georeferenced semantic maps (e.g. Fig.~\ref{fig:color-segmentation}), enabling large-scale, precise spatial analysis. 
The map size, however, increases significantly when applying deep neural networks on large orthomosaics. Instead of relying on expensive computation, we propose an affordable solution that trains and infers on small tiles split from a large map. We present a registration algorithm to merge semantic objects from multiple tiles during inference. 

\begin{figure}[h]
\centering
\vspace{6pt}
\includegraphics[width=0.40\textwidth]{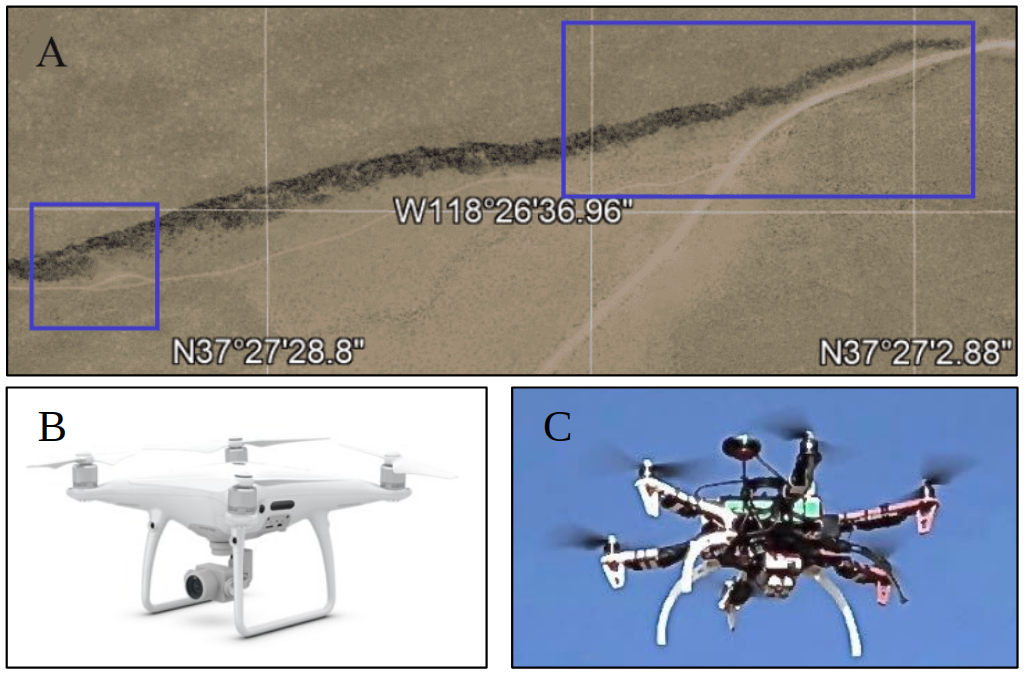}
\caption{Areas of study and UAVs. (A) Areas of study. The top-right rectangle: experiment I; the bottom-left rectangle: experiment II. (B) DJI Phantom 4 Pro with an RGB camera. (C) Hexrotor with a MicaSense RedEge MX.}
\label{fig:area-study}
\end{figure}

We applied the UAS-SfM-DL system to two experiments analyzing rock traits along a tectonic fault scarp in Bishop, California (Fig.~\ref{fig:area-study}). In the first experiment, we deployed a DJI Phantom 4 Pro with an RGB camera to demonstrate proof of concept, and inspected how much elevation information improved instance segmentation from RGB channels. In the second experiment, we equipped a hexrotor with a multispectral camera (MicaSense RedEdge MX) that can capture spectral imagery from five bands: red (R), green (G), blue (B), red edge (RE), and near infrared (NIR). The orthomosaics of the five channels and the DEM acquired from SfM were used to train a deep neural network to detect and segment individual rocks. We compared the neural network inference performance for different input channel combinations. This UAS-SfM-DL system presents a new way to automatically characterize surface processes on fault scarps.

\section{Related Work}

\subsection{Unpiloted Aircraft Systems and Structure from Motion (UAS-SfM)}
Structure from motion originated in the computer vision community and has become popular with the utilization of bundle adjustment optimization \cite{snavely2010scene}. With the recent availability of low cost UAS and software such as Agisoft \cite{agisoft}, UAS-SfM have widely been used in physical geography \cite{westoby2012structure} ranging from coastal environments \cite{mancini2013using}, to Antarctic moss beds \cite{lucieer2011unmanned}, to fault scarps \cite{donnellan2018geodetic, johnson2014rapid}. However, geomorphological analysis of products from UAS-SfM still depends a lot on interpretive models carefully designed by experts \cite{cook2017evaluation, bunds2018three}.

\subsection{Deep Learning in Close-range Aerial Imagery Processing}
The advances in deep learning have facilitated the development of visual perception models deployed aboard UAS as well as ground vehicles in various applications like weed classification \cite{hung2014feature}, car detection \cite{ammour2017deep}, and fruit counting \cite{chen2017counting}. However, previous work largely targets camera perspective imagery as the input to the deep learning models. Although feature tracking algorithms \cite{liu2019monocular} can reconstruct objects of interest, they lack ground control points (GCPs) to globally correct geographic distortions, which is an essential step in geological and surveying applications. Deep learning has also been used to process 3D information, such as LiDAR point clouds. For example, point cloud data generated from scanning trees were processed by fully connected layers for tree classification \cite{guan2015deep}. 

\section{System Description}
The workflow of our UAS-SfM-DL pipeline is shown in Fig.~\ref{fig:pipeline}. Although each component in the pipeline is not new, we focus on system integration and solutions to practical implementation issues involved in this fault scarp application in geomorphological analysis. Additionally, we present the pipeline from a high-level perspective to avoid curbing the generalization to other potential applications. 

Aerial imagery from UAS along with GCPs are processed using SfM algorithms to reconstruct a high-resolution study site. Geologists annotate features of interest, such as rock boundaries, from a portion of the RGB orthomosaics. We train deep neural networks on annotated images, and carry out inference on unlabelled images from the entire site. Segmented objects inferred by the deep learning models are post-processed by geometric structural analysis and shape descriptors to estimate properties such as rock size, roundness, and orientation. We generate semantic maps by combining the post-processed inference results with georeferenced metadata. Statistics describing the distribution of rock traits are acquired from the semantic maps, and they can be used in statistical descriptions for future geomorphological studies.

\subsection{UAS-SfM} 

Georeferenced aerial imagery collected from UAS and GCPs measured from differential GPS devices are processed by bundle-adjustment-based SfM algorithms~\cite{triggs1999bundle}, which produce precise georeferenced products including point clouds, DEM, DSM, and orthomosaics for each spectral band. In practice, the orthomosaics and DEM are formatted as GeoTiff files with metadata such as global coordinates, projection type, and ground resolution. These data give us access to global coordinates of each 3D point or pixel in the models, which enables succeeding spatial analysis of semantic features from deep learning.

\subsection{Deep Learning}

\begin{figure}[t]
\centering
\vspace{6pt}
\includegraphics[width=0.40\textwidth]{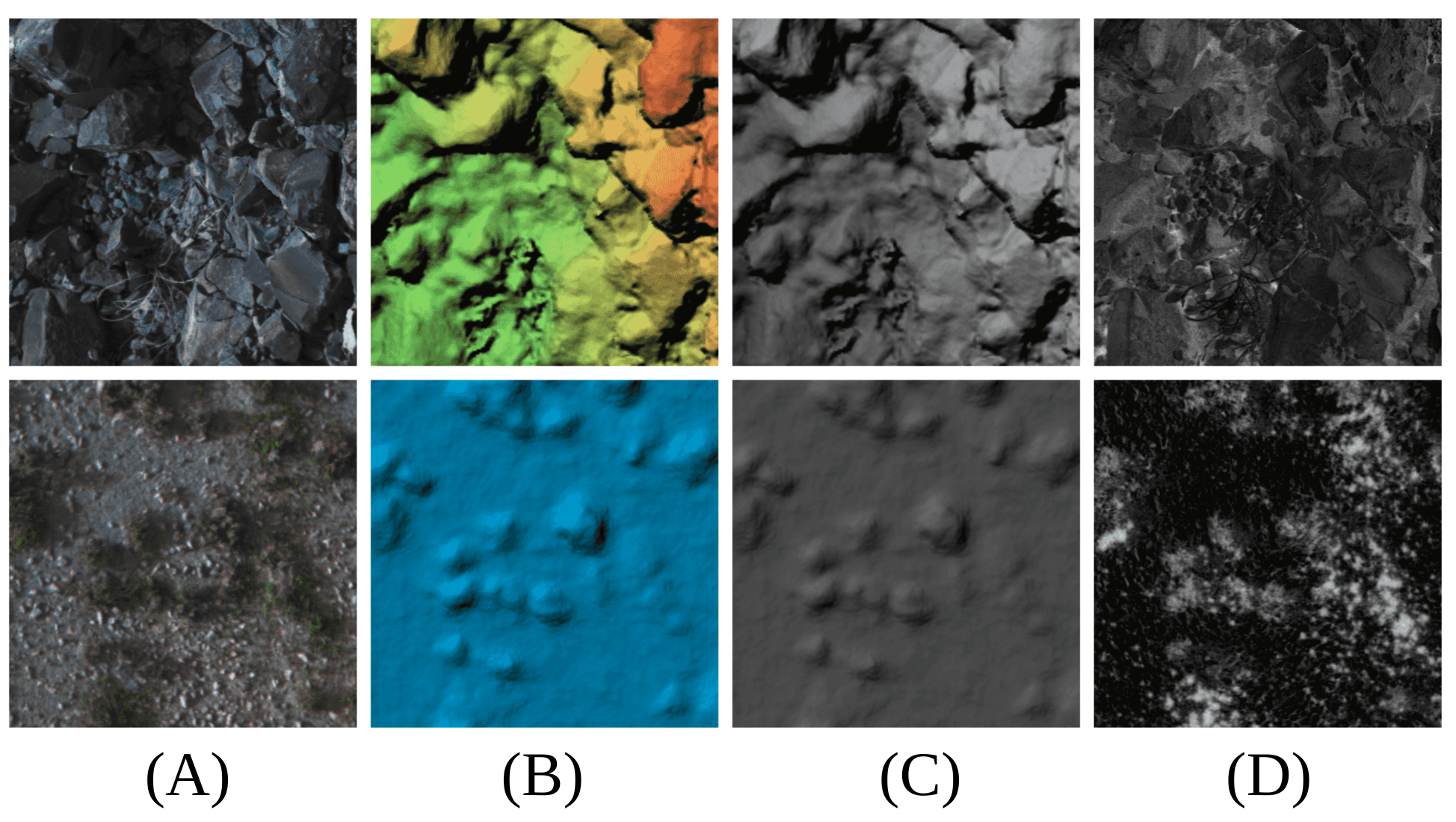}
\caption{Spectral and DEM representations. All tiles on each row are of the same area. 
Column (A) RGB orthomosaic tiles. Column (B) Colormap elevation tiles. Column (C) Relative elevation tiles. Column (D) NDVI orthomosaic tiles.}
\label{fig:rgb_dem_ndvi}
\end{figure}

We use deep neural networks to automatically extract semantic information in large-scale georeferenced models produced from SfM. Deep learning accommodates versatile, selectable inputs such as RGB orthomosaics, DEMs, and other spectral orthomosaics from multispectral cameras. The selection of neural network architecture depends on the features and distributions of objects of interest. We use the Mask R-CNN deep neural network architecture \cite{he2017mask} for this study because it generates both bounding boxes and the corresponding masks for object instances (rocks). Faster R-CNN  \cite{ren2015faster} with a large number of Regions of Interest is adopted for the object detection branch in the Mask R-CNN because of the dense spatial distribution of the rocks at this fault scarp. Additionally, because of the large range of rock sizes (major-axis lengths ranging between 0.2-3.6 meters), we select the Pyramid Feature Network (PFN) \cite{lin2017feature} as the backbone of the Faster R-CNN. PFN's multi-scale, pyramidal hierarchy of anchor generation mechanism is suitable for object detection in such a setting. For other potential studies, segmentation neural networks such as U-Net~\cite{ronneberger2015u} can be used when a goal is a pixel-level segmentation. Region proposal networks such as Faster R-CNN \cite{ren2015faster} can detect individual objects with bounding boxes, when instance segmentation is not demanded.

While it is common to scale spectral orthomosaics by intensity values, directly scaling DEM by elevation range values will overshadow the rock information in the scaled space. This is because the fault scarp slope elevation (20-30 meters) is greater than the height of most of the rocks (0.2-3 meters). Instead of compressing elevation into one single channel, we use a colormap (3 channels) to encode the DEM:
\begin{equation}
    D_{rgb}=c(s(h))
\end{equation}
where $h$ is absolute elevation, $s$ is a linear scale function $s:[h_{min}, h_{max}]\rightarrow[0,1]$, and $c$ is a colormap function mapping scaled value to RGB color domain $c:[0,1]\rightarrow\{r, g, b\}$. In doing so, we capture the rock elevation in the three channels so that the rock height will not be overshadowed. An example of jet colormap is shown in Fig.~\ref{fig:rgb_dem_ndvi}(B). Because colormap elevation represents absolute elevation, one concern is that deep learning networks may be constrained to learn features only from a certain range of absolute elevation. This issue may become more serious in this study because the number of rocks on the fault scarp is greater than the number of rocks on the lower-side hanging wall.

Apart from colormap elevation, we present another elevation representation that preserves local, relative elevation: 
\begin{equation}
    D=g(D_{rgb})=\frac{D_r}{3} + \frac{D_g}{3} + \frac{D_b}{3}
\end{equation}
where $D_r$, $D_g$, and $D_b$ are three channels from $D_{rgb}$. The relative elevation representation (Fig.~\ref{fig:rgb_dem_ndvi}(C)) is superior in the sense that it can reflect local elevation such that deep neural networks can attain generalization to detect and segment rocks on any elevations. One potential concern of relative elevation is that the mapping $g(c(s(\cdot)))$ is noncontinuous on the absolute elevation domain, which may result in high-frequency noises in relative elevation tiles. However, our experiments in the next section show that such high-frequency noises will not cause problems for deep neural networks.

\begin{figure}
\centering
\vspace{6pt}
\includegraphics[width=0.48 \textwidth]{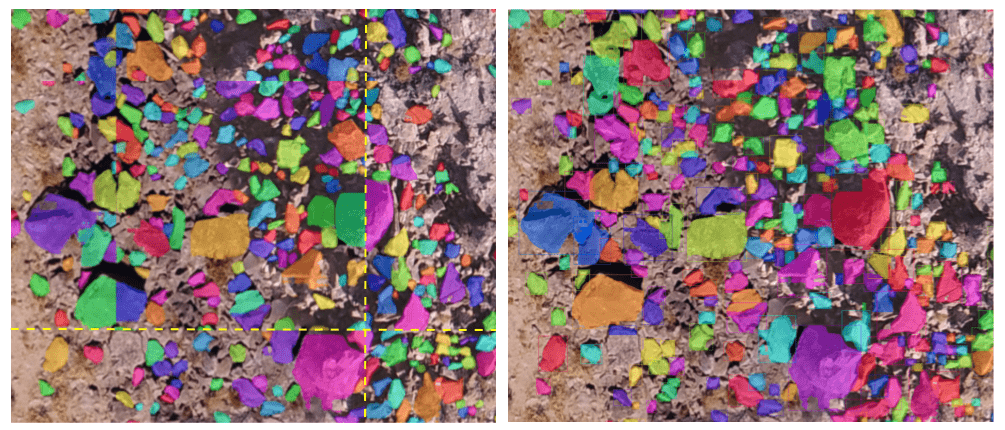}
\caption{Rock registration at tile boundaries. (Left) Rocks detected by the neural network at the edges of each inference are not merged; (right) any two rocks at the edges of each inference are merged and registered as one if they belong to the same instance.}
\label{fig:rock-registration}
\end{figure}

{\small 
\begin{table} 
\resizebox{\columnwidth}{!}{%
\begin{tabular}{l}
\hline
\textbf{Algorithm 1} Rock Registration                                                                   \\ \hline
\begin{tabular}[c]{@{}l@{}}\textbf{input}: orthomosaics, neural\_network\\ \textbf{output}: registered\_rocks\end{tabular} \\ \hline
1. tiles = overlap\_split(orthomosaics)                                                                  \\ 
2. rocks = project(neural\_network(tiles))                                                                    \\ 
3. registered\_rock = Empty\_list                                                                               \\
4. \textbf{for} rock \textbf{in} rocks:                                                                                 \\ 
\hspace{7mm}\textbf{if} rock.bbox is on its own tile edges:                                                             \\ 
\hspace{11mm} id = check\_bbox\_overlap(rock.bbox, registered\_rocks)                                       \\
\hspace{11mm} \textbf{if} id $\neq$ None:                                                                       \\
\hspace{15mm} \textbf{if} check\_mask\_overlap(rock.mask, registered\_rocks[id].mask)                           \\
\hspace{18mm} > threshold:                                                                             \\
\hspace{19mm} merge(registered\_rocks[id], rock)                                                       \\
\hspace{19mm} \textbf{continue}                                                       \\
\hspace{7mm} register(registered\_rocks, rock)                                                       \\ \hline
Comments: \\
\multicolumn{1}{p{8.5cm}}{\textbf{overlap\_split} splits orthomosaics into 400x400 pixel tiles with 10-pixel overlap at four edges} \\
\multicolumn{1}{p{8.5cm}}{\textbf{project} projects local bounding boxes with pixel coordinates to global coordinates } \\
\multicolumn{1}{p{8.5cm}}{\textbf{check\_bbox\_overlap} returns None if there is no registered rocks having bounding box overlap with the rock, otherwise returns the overlapped registered rock's id}\\
\multicolumn{1}{p{8.5cm}}{\textbf{check\_mask\_overlap} returns the intersection of two masks}\\
\multicolumn{1}{p{8.5cm}}{\textbf{merge} merges the bounding boxes and masks to the registered rock}\\ 
\multicolumn{1}{p{8.5cm}}{\textbf{register} registers a new rock}\\ \hline
\end{tabular}
}
\end{table}
}

\subsection{Tiling and Registration}
The orthomosaics of survey sites produced by SfM are of high resolution and large scale (for our first fault scarp study, 2 cm/pixel, 25664x10589 pixels). Directly working with high-resolution images for neural network training or inference is computationally challenging because it places a high demand on GPU RAM. To address this limitation, we split the orthomosaics into smaller tiles (400x400 pixels). We select a subset of tiles and annotate rocks as bounding polygons in LabelMe~\cite{russell2008labelme}, then divide them into a training dataset and a testing dataset. We augment the training dataset with a combination of random left-right flipping, top-down flipping, rotation, and zooming-in (cropping) and zooming-out. 

Splitting the orthomosaics into tiles causes some rocks to be divided. As a result, rocks detected at the edges of tiles may belong to one single instance and risk getting treated as several smaller rocks as shown in Fig.~\ref{fig:rock-registration}(right). To address this problem, we split the orthomosaics into 400x400 pixel tiles with 10-pixel overlaps on four edges when carrying out inference. A registration scheme is applied to merge objects detected at the edges of tiles if they are from a single instance (Algorithm I). We determine that two georeferenced rock instances within the 10-pixel region of overlap are the same instance by checking if the intersection of their masks is greater than a threshold. If true, the two rock instances are merged and registered as one instance (Fig.~\ref{fig:rock-registration}(left)). To keep the comparison in \textbf{check\_bbox\_overlap} from becoming quadratic with rock numbers, we utilize the spatial relation and only compare rock overlaps for rocks on the 10-pixel overlap zones of four neighbor tiles. 

\subsection{Post Processing}
Post processing is necessary to identify the rocks on the fault scarp, and to compensate for errors resulting from deep neural networks. Because only rocks near the fault scarp are of interest for tectonic study, we need to clear away rocks detected on the perimeter. As the fault scarp has a steep slope, we first estimate the gradient of the DEM \cite{horn1981hill}, then denoise it with Gaussian filtering. From the smoothed slope map, the fault scarp is identified by the slope above a certain threshold and further processed through morphological transformations such as opening, dilation, erosion, etc. 

\begin{figure}
\centering
\includegraphics[width=0.48\textwidth]{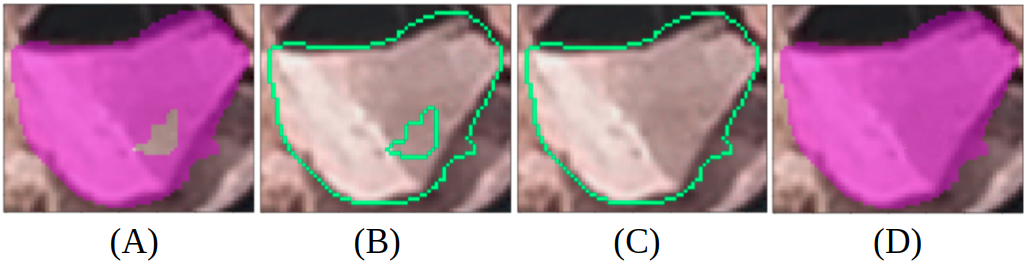}
\caption{Contour analysis to compensate for instability of segmentation prediction. (A) Mask prediction from Mask R-CNN. (B) Contours by topological structure analysis. (C) The largest contour kept to approximate outline of the rock. (D) Filtered mask.}
\label{fig:topological-filtering}
\end{figure}

The segmentation sub-network has stochastic errors, with an example shown in Fig.~\ref{fig:topological-filtering}(A) where a hole is present in the segmented rock body. We assume that there are no rocks with torus topology on the fault scarp, and use topological structure analysis to remove such artifacts \cite{suzuki1985topological}. The contours of rocks are generated, with the largest or most exterior contours retained as the outlines of the rocks. The rock sizes are approximated by the number of pixels in the largest contours. Note that georeferenced models enable us to associate rock size in pixels and rock size in meters, which is more accurate than the association from perspective 2D camera photos where object scale is ambiguous with depth.

\begin{figure} 
\centering
\vspace{6pt}
\includegraphics[width=0.4\textwidth]{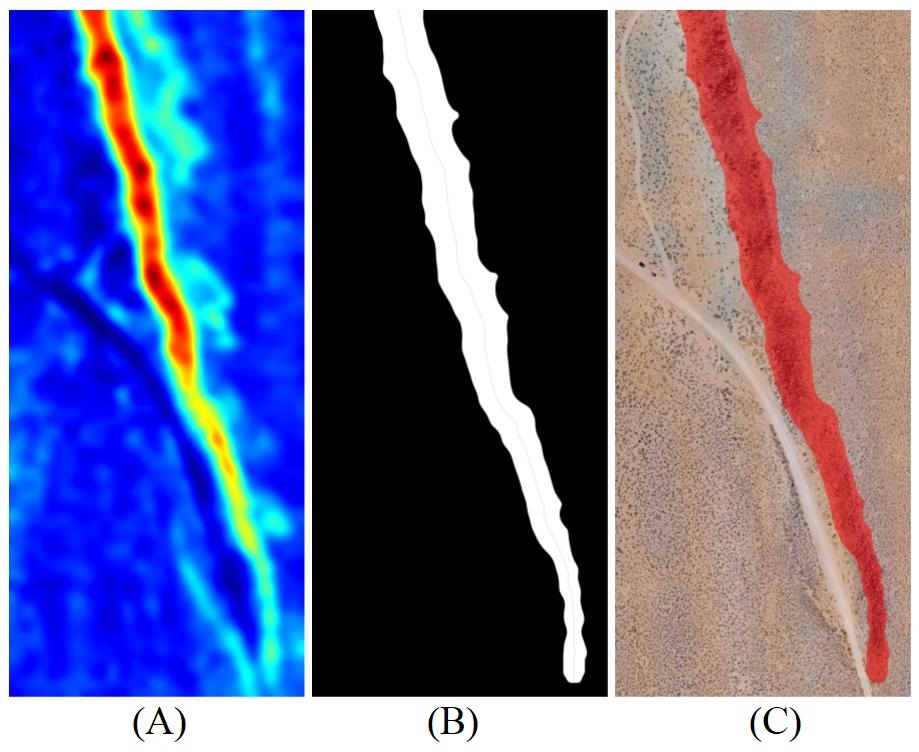}
\caption{Identifying the fault scarp. (A) Slope map. (B) Fault scarp contour. (C) Overlap with RGB orthomosaics.}
\label{fig:identifying-fault-scarp}
\end{figure}

\begin{figure}[]
\centering
\includegraphics[width=0.45\textwidth]{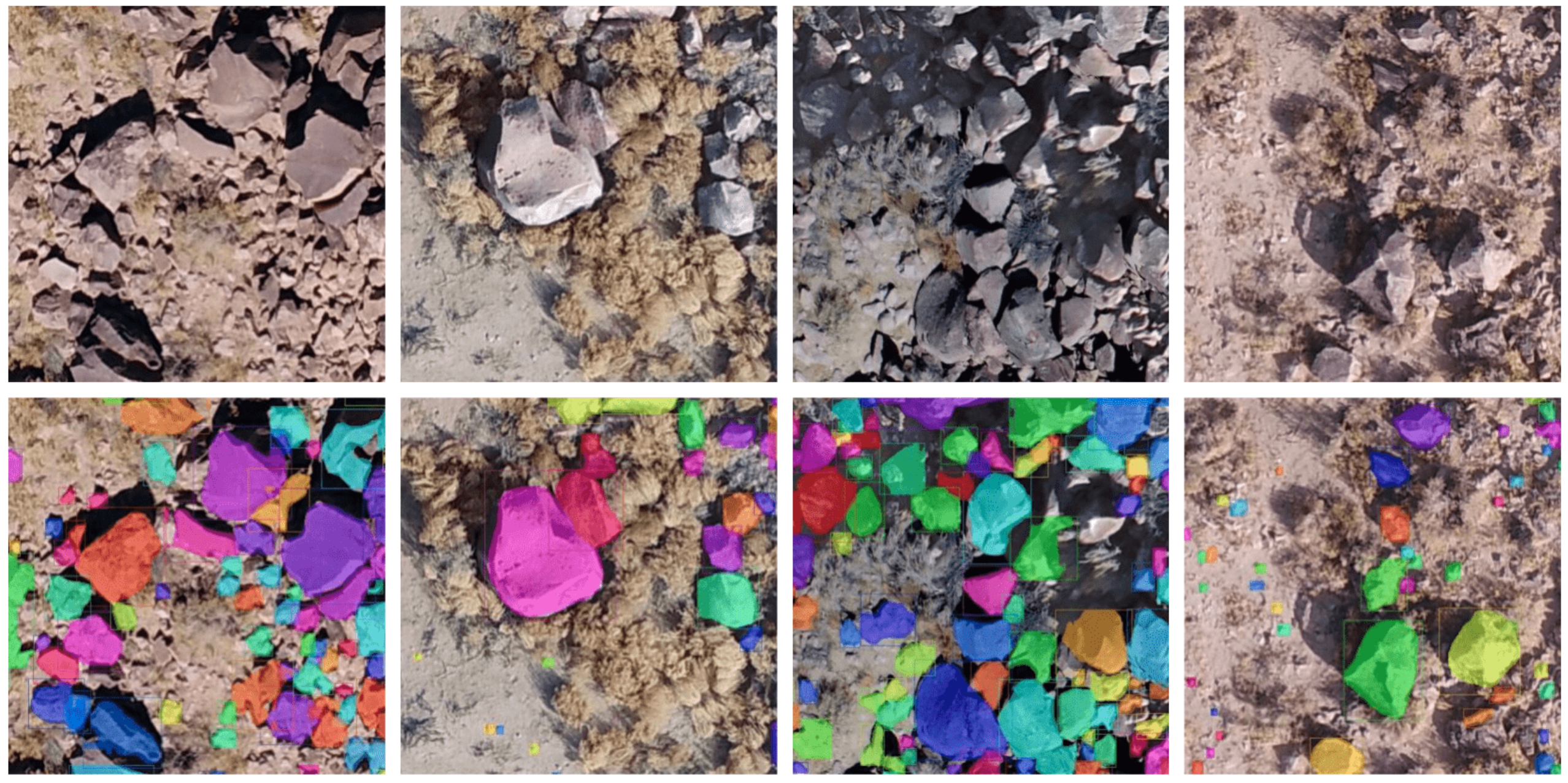}
\caption{Prediction of Mask R-CNN on test dataset sample tiles. Colors are randomly selected to distinguish rocks.}
\label{fig:tiles}
\end{figure}

{\small
\begin{table*}[h!]
\centering
\vspace{10pt}
\caption{Experiment I: Neural Network Inference Results*}
\begin{adjustbox}{width=1.8\columnwidth,center}
\begin{tabular}{|c|c|l|l|l|l|l|l|l|l|l|l|l|l|l|}
\hline
\multicolumn{3}{|l|}{\multirow{2}{*}{}} & \multicolumn{6}{c|}{\begin{tabular}[c]{@{}c@{}}Detection\\ (Bounding Box)\end{tabular}}                                                                                     & \multicolumn{6}{c|}{\begin{tabular}[c]{@{}c@{}}Segmentation\\ (Mask)\end{tabular}}                                                                                          \\ \cline{4-15} 
\multicolumn{3}{|l|}{}                  & \multicolumn{1}{c|}{$AP_1$} & \multicolumn{1}{c|}{$AP_2$} & \multicolumn{1}{c|}{$AP_3$} & \multicolumn{1}{c|}{$AP_4$} & \multicolumn{1}{c|}{$AR_1$} & \multicolumn{1}{c|}{$AR_2$} & \multicolumn{1}{c|}{$AP_1$} & \multicolumn{1}{c|}{$AP_2$} & \multicolumn{1}{c|}{$AP_3$} & \multicolumn{1}{c|}{$AP_4$} & \multicolumn{1}{c|}{$AR_1$} & \multicolumn{1}{c|}{$AR_2$} \\ \hline
\multirow{3}{*}{TL}    & 1  & RGB       & 21.7                       & 46.5                       & 17.0                       & 36.9                       & 31.4                       & 48.3                       & 20.8                       & 46.5                       & 14.6                       & 48.6                       & 29.8                       & 51.7                       \\ \cline{2-15} 
                       & 2  & RGB+DEM3  & 22.7                       & 45.3                       & 19.4                       & 51.1                       & 32.1                       & \textbf{58.3}              & 21.2                       & 45.7                       & \textbf{16.4}              & 50.7                       & 30.3                       & 53.3                       \\ \cline{2-15} 
                       & 3  & RGB+DEM1  & \textbf{23.1}              & \textbf{46.9}              & \textbf{20.5}              & \textbf{53.9}              & \textbf{32.6}              & \textbf{58.3}              & \textbf{25.1}              & \textbf{47.4}              & 16.2                       & \textbf{61.0}              & \textbf{30.8}              & \textbf{65.0}              \\ \hline
\multirow{3}{*}{NTL}   & 4  & RGB       & 21.2                       & \textbf{46.2}              & 16.5                       & \textbf{42.9}              & 29.9                       & 43.3                       & 21.2                       & 45.7                       & 16.3                       & 42.9                       & 29.3                       & 50.0                       \\ \cline{2-15} 
                       & 5  & RGB+DEM3  & 20.5                       & 42.6                       & \textbf{16.6}              & 40.2                       & 29.9                       & 45.0                       & 19.4                       & 42.4                       & 14.6                       & 48.6                       & 28.1                       & 53.3                       \\ \cline{2-15} 
                       & 6  & RGB+DEM1  & \textbf{21.4}              & 45.9                       & 16.1                       & 40.6                       & \textbf{30.9}              & \textbf{48.3}              & \textbf{21.6}              & \textbf{46.6}              & \textbf{16.5}              & \textbf{49.9}              & \textbf{31.0}              & \textbf{56.7}              \\ \hline

\multicolumn{15}{l}{\begin{tabular}[c]{@{}l@{}}* TL: transfer learning, NTL: no transfer learning, IoU: intersection over union, $AP_1$: average precision (\%) with \\IoU=0.5:0.95, $AP_2$: average precision (\%) with IoU=0.5, $AP_3$: average precision (\%) with IoU=0.75, $AP_4$: average \\precision (\%) with IoU=0.5:0.95 for large objects, $AR_1$: average recall (\%) with IoU=0.50:0.95, \\$AR_2$: average recall (\%) with IoU=0.50:0.95 for large objects\end{tabular}}
\end{tabular}
\end{adjustbox}
\label{table:c3}
\vspace{-5mm}
\end{table*}
}

\section{Experiments}
In this section, we discuss the results of our pipeline in two experiments at the Volcanic Tablelands, which is a faulted plateau of approximately 150-meter-thick welded Bishop Tuff (760 ka) at the north end of Owens Valley near Bishop, California \cite{ferrill2016observations}. 

\subsection{DJI Phantom 4 Pro and RGB Camera}
The study area for the first experiment is shown in the top-right rectangle in Fig.~\ref{fig:area-study}(A). One goal of this experiment is to demonstrate the proof of concept by going through implementation details in the presented pipeline and obtaining the rock-trait histograms. We conducted surveys of the Volcanic Tablelands with a DJI Phantom 4 Pro in March 2018. A grid flight pattern was implemented with \text{66\%} image overlap and a 90° (nadir) camera angle. The flight altitude varied between 70-100 meters above the ground level because the height of the fault scarp slope is around 30 meters. The onboard RGB camera had an 84° field of view and 5472x3648 resolution.

We used Agisoft~\cite{agisoft} for SfM and produced a DEM and an RGB orthomosaic (25664x10589 pixels, 2 cm/pixel). The orthomosaics were split into 400x400 tiles. It took about 65 work hours for a human expert to annotate 67 tiles with 4095 rocks (49 tiles for training, and 18 for testing). We trained Mask R-CNN (ResNet-50 backbone) on an Nvidia RTX 2080 Ti, and the inference results on the test dataset are shown in Table~\ref{table:c3}. 

For transfer learning, the neural network weights were initialized with training results from COCO 2017, except that the first and the last layers were initialized from a uniform distribution. When comparing trials between TL and NTL (Table~\ref{table:c3}), we found the neural networks benefited from transfer learning. Both colormap elevation (DEM3) and relative elevation (DEM1) improved neural network performance from RGB orthomosaics in the case of transfer learning. DEM3 assisted the neural network performance in transfer learning but decreased in non-transfer learning. Considering key performances ($AP_1$, $AR_1$ for both detection and segmentation), RGB+DEM1 outperformed others in both transfer learning and non-transfer learning. 

Once neural network inference was conducted on all tiles from the study area, we carried out post-processing to obtain rock-trait histograms at the fault scarp. We removed outliers on the perimeter and identified the boundaries of the fault scarp from the slope map utilizing terrain gradient. The slope map and the fault scarp contour are shown in Fig.~\ref{fig:identifying-fault-scarp}. The enclosed rocks were then estimated by the refined masks from topological structure analysis. The results of the filtered rock instances at the fault scarp of this study area are shown in Fig.~\ref{fig:color-segmentation}. Some predictions randomly selected from the test dataset are shown in Fig.~\ref{fig:tiles}.

\begin{figure}[]
\centering
\includegraphics[width=0.45\textwidth]{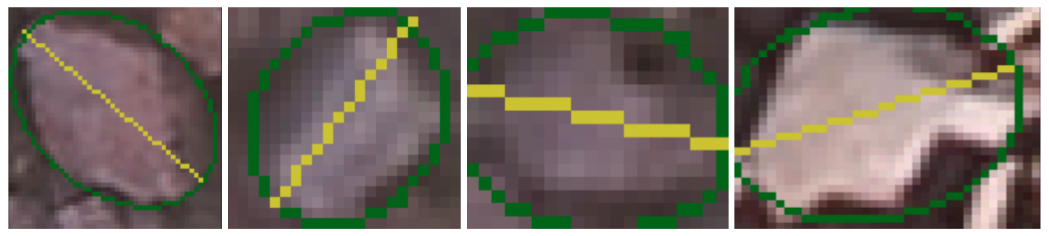}
\caption{Ellipse fitting and estimation of major orientations from the inferred rock boundaries. Yellow line indicates major axis.}
\label{fig:ellipse-fitting}
\end{figure}

\begin{figure}[t]
\centering
\includegraphics[width=0.48\textwidth]{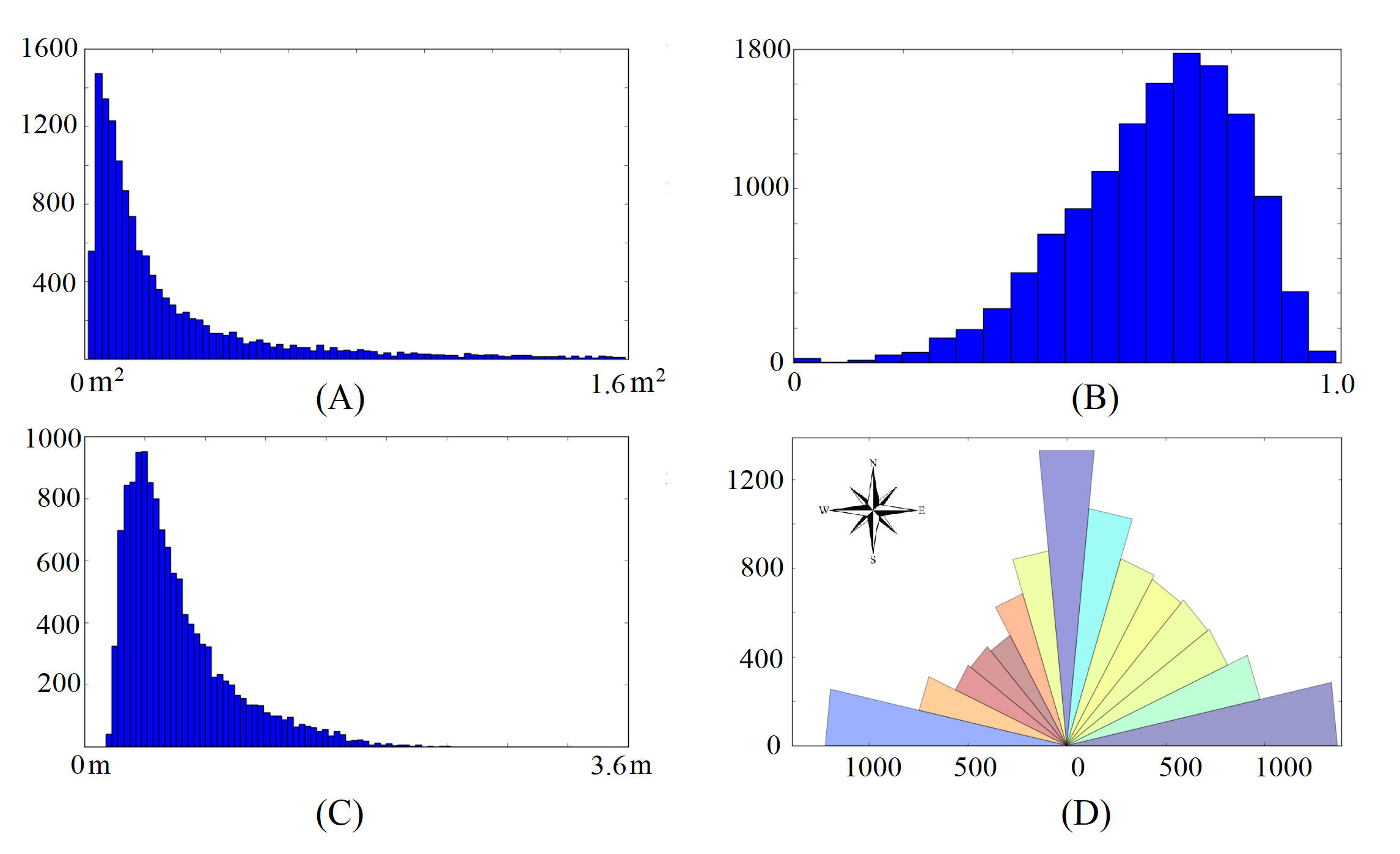}
\caption{Histograms of rock traits. (A) Rock size histogram of the fault scarp. The rock sizes are approximated by the area of the refined masks. The horizontal axis is rock area in $meters^2$, and the vertical axis is the number of rocks. (B) Rock eccentricity histogram of the fault scarp. (C) Rock major-axis length histogram on the fault scarp. The major-axis length is $L=2a$, where $a$ is the semimajor axis of the fitting ellipse $\frac{x^2}{a^2}+\frac{y^2}{b^2}=1$. (D) Polar histogram of major-axis orientation at the fault scarp.}
\label{fig:rock-size-hist}
\end{figure}

Lastly, we used the georeferenced rock boundary information to estimate rock diameter, roundness, and orientation. To approximate roundness and major orientations of rocks, we fitted the refined masks with ellipses as shown in Fig.~\ref{fig:ellipse-fitting}. We used ellipse eccentricity to describe rock roundness. The orientations of rocks were parameterized by orientations of the ellipses' major axes. Fig.~\ref{fig:rock-size-hist} shows rock-trait histograms of the fault scarp. In this experiment, 12,682 rocks were detected along the fault scarp in the study area, which only has an area of 513 meters by 212 meters. 

\begin{figure}[h!] 
\centering
\vspace{6pt}
\includegraphics[width=0.4\textwidth]{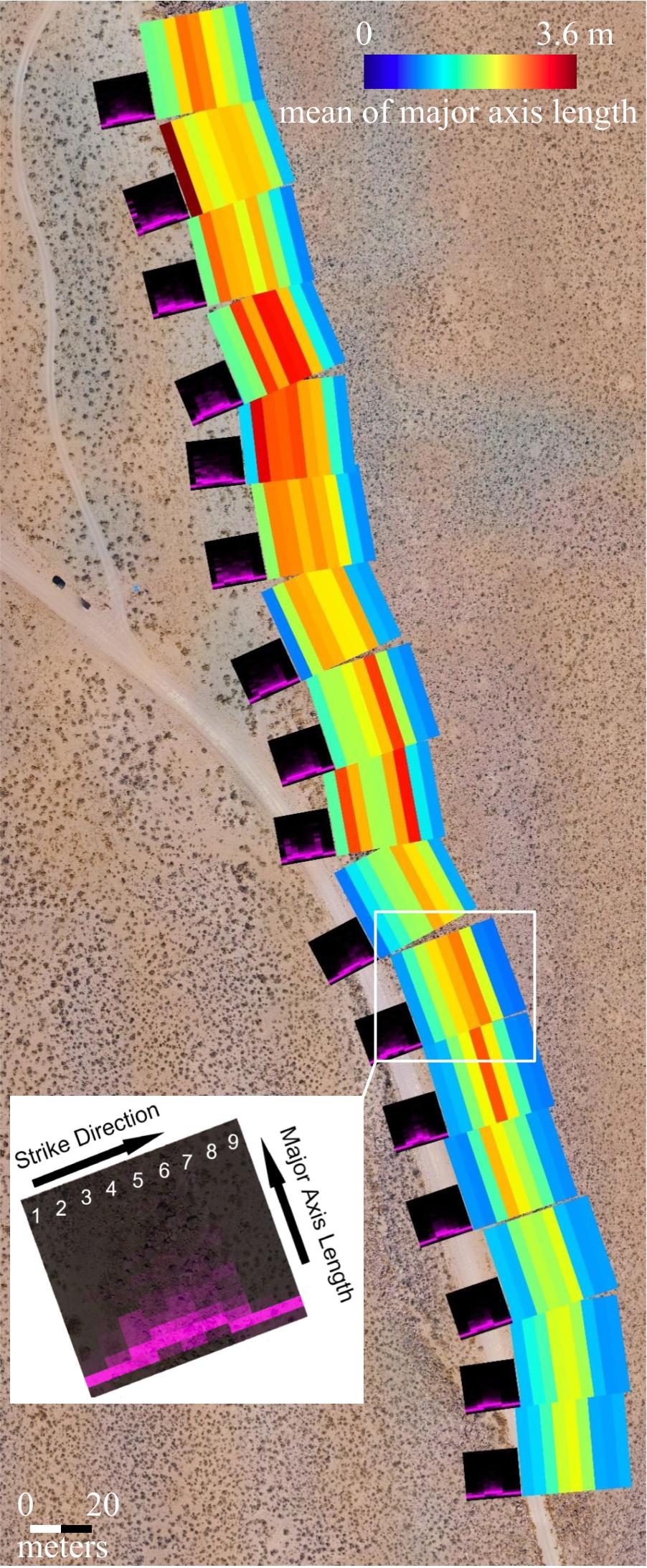}
\caption{Detailed rock diameter (major-axis length) histogram colormap for the study area. The colormap on the fault scarp shows the mean rock diameter for each box along the scarp. The smaller black plots at the left show detailed histograms of rock diameter with transparency indicating lower spatial density.}
\label{fig:rock-size-hist-area}
\end{figure}

We are not only interested in the distribution of rock traits along the length of the scarp, but along the cross strike as well, which is the perpendicular direction of the fault scarp (Fig.~\ref{fig:rock-size-hist-area}). We computed the skeleton of the fault scarp contour to acquire the middle spline \cite{zhang1984fast}, then selected a subsection of the spline and approximated it with a straight line by linear regression. We considered the normal vector of the straight line as the cross strike direction. The whole fault scarp is divided into 16 areas that are lying in the center of the spline. Each area is divided into 9 boxes along the cross strike direction. Within each box, there are 20 bins representing the normalized rock diameter histogram in the true box. From the bottom (south) to top (north) in each area, the histogram axis (major-axis length) varies from 0 to 3.6 meters.

\subsection{Hexrotor and Multispectral Camera}
In this experiment, we focused on examining the presented pipeline on different combinations of inputs including multispectral orthomosaics, color elevation, and relative elevation. The study area for this experiment is shown in the bottom-left rectangle in Fig.~\ref{fig:area-study}(A). We equipped a hexrotor with an RTK GPS and a straight-down multispectral camera (MicaSense RedEdge MX) that can capture spectral bands of blue (465-485 nm), green (550-570 nm), red (663-673 nm), red edge (712-722 nm), and near infrared (820-860 nm). We deployed a lawnmower pattern with a flight altitude of 30-50 meters above the fault scarp (the height of the fault scarp slope is about 20 meters) and collected synchronized multispectral imagery in April 2019. Agisoft ~\cite{agisoft} was used for SfM and produced a DEM and five spectral orthomosaics (9258x8694 pixels, 2 cm/pixel). We split the spectral orthomosaics and DEM into 400x400 tiles with 10-pixel size overlap at four edges. It took about 50 work hours for a human expert to annotate 37 tiles for training and 18 tiles for testing (2541 rocks annotated in total). We trained Mask R-CNN (ResNet-50 backbone) on an Nvidia RTX 2080 Ti. 

The inference results on the test dataset are shown in Table~\ref{table:bishop}. We consider RGB results (trials 1 and 8) as the baseline and discuss the neural network performances with different input combinations. Because there are several different performance metrics in the table and neural network prediction can be noisy, we emphasize four key metrics - $AP_1$ and $AR_1$ from detection and segmentation.

{\small
\begin{table*}[h!]
\centering
\vspace{10pt}
\caption{Experiment II: Neural Network Inference Result*}
\begin{adjustbox}{width=1.8\columnwidth,center}
\begin{tabular}{|c|c|l|l|l|l|l|l|l|l|l|l|l|l|l|}
\hline
\multicolumn{3}{|l|}{\multirow{2}{*}{}}     & \multicolumn{6}{c|}{\begin{tabular}[c]{@{}c@{}}Detection\\ (Bounding Box)\end{tabular}}                                                                                     & \multicolumn{6}{c|}{\begin{tabular}[c]{@{}c@{}}Segmentation\\ (Mask)\end{tabular}}                                                                                          \\ \cline{4-15} 
\multicolumn{3}{|l|}{}                  & \multicolumn{1}{c|}{$AP_1$} & \multicolumn{1}{c|}{$AP_2$} & \multicolumn{1}{c|}{$AP_3$} & \multicolumn{1}{c|}{$AP_4$} & \multicolumn{1}{c|}{$AR_1$} & \multicolumn{1}{c|}{$AR_2$} & \multicolumn{1}{c|}{$AP_1$} & \multicolumn{1}{c|}{$AP_2$} & \multicolumn{1}{c|}{$AP_3$} & \multicolumn{1}{c|}{$AP_4$} & \multicolumn{1}{c|}{$AR_1$} & \multicolumn{1}{c|}{$AR_2$} \\ \hline
\multirow{7}{*}{TL}  & 1  & RGB             & 25.7                       & 60.7                       & 17.2                       & 51.9                       & 37.0                       & 57.0                       & 23.4                       & 51.8                       & 15.3                       & 52.9                       & 27.8                       & 55.0                       \\ \cline{2-15} 
                     & 2  & DEM1            & 4.9                       & 15.3                       & 2.3                       & 39.5                       & 11.2                       & 58.0                       & 4.3                       & 12.5                       & 2.0                       & 45.5                       & 9.9                       & 53.0                       \\ \cline{2-15} 
                     & 3  & RGB+DEM3        & 28.4                       & 64.0                       & 19.7                       & 64.9                       & 39.5                       & 69.0                       & 26.6                       & 61.6                       & 18.7                       & 61.1                       & 36.3                       & \textbf{62.0}              \\ \cline{2-15} 
                     & 4  & RGB+DEM1        & 28.8                       & \textbf{64.7}              & 22.1                       & 59.1                       & 40.1                       & 63.0                       & 26.8                       & 61.8                       & 19.1                       & 54.5                       & 36.3                       & 56.0                       \\ \cline{2-15} 
                     & 5  & RGB+RE+NIR      & 28.5                       & 61.7                       & 22.2                       & 58.8                       & 38.8                       & 61.0                       & 25.8                       & 57.2                       & 20.1                       & 57.1                       & 35.1                       & 58.0                       \\ \cline{2-15} 
                     & 6  & RGB+RE+NIR+DEM3 & 29.6                       & 63.4                       & \textbf{23.7}              & \textbf{60.2}              & 39.1                       & \textbf{64.0}              & \textbf{29.0}              & \textbf{62.0}              & \textbf{22.1}              & \textbf{54.9}              & 37.7                       & 56.0                       \\ \cline{2-15} 
                     & 7  & RGB+RE+NIR+DEM1 & \textbf{29.7}              & \textbf{64.7}              & \textbf{23.7}              & 58.3                       & \textbf{40.6}              & 63.0                       & 28.3                       & 61.6                       & 21.0                       & 54.4                       & \textbf{38.2}              & 56.0                       \\ \hline
\multirow{7}{*}{NTL} & 8  & RGB             & 24.9                       & 60.4                       & 16.1                       & 40.2                       & 34.2                       & 45.0                       & 22.5                       & 55.9                       & 14.3                       & 43.5                       & 31.2                       & 45.0                       \\ \cline{2-15} 
                     & 9  & DEM1            & 3.8                       & 11.5                       & 1.8                       & 31.7                       & 8.7                       & 45.0                       & 3.3                       & 9.8                       & 1.1                       & 31.1                       & 7.3                       & 36.0                       \\ \cline{2-15} 
                     & 10 & RGB+DEM3        & 26.0                       & 62.3                       & 17.9                       & \textbf{45.0}              & 36.0                       & 55.0                       & 24.1                       & 57.4                       & 16.3                       & \textbf{51.6}              & 33.6                       & \textbf{54.0}              \\ \cline{2-15} 
                     & 11 & RGB+DEM1        & 27.1                       & 63.8                       & \textbf{19.0}              & 36.6                       & 38.5                       & 41.0                       & 25.2                       & 62.0                       & 16.2                       & 44.4                       & 35.5                       & 45.0                       \\ \cline{2-15} 
                     & 12 & RGB+RE+NIR      & 27.6                       & 63.5                       & 18.7                       & 41.5                       & \textbf{39.3}              & \textbf{47.0}              & 23.5                       & 59.6                       & 14.6                       & 49.1                       & 34.2                       & 51.0                       \\ \cline{2-15} 
                     & 13 & RGB+RE+NIR+DEM3 & 26.9                       & 63.5                       & 17.0                       & 32.1                       & 38.8                       & 37.0                       & 23.2                       & 59.3                       & 14.3                       & 30.6                       & 34.0                       & 30.0                       \\ \cline{2-15} 
                     & 14 & RGB+RE+NIR+DEM1 & \textbf{28.0}              & \textbf{65.8}              & 17.8                       & 40.0                       & \textbf{39.3}              & 41.0                       & \textbf{25.8}              & \textbf{62.2}              & \textbf{16.7}              & 47.7                       & \textbf{35.7}              & 49.0                       \\ \hline
\multicolumn{15}{l}{\begin{tabular}[c]{@{}l@{}}* Refer to the table note in Table~\ref{table:c3}\end{tabular}}
\end{tabular}
\end{adjustbox}
\label{table:bishop}
\vspace{-5mm}
\end{table*}
}

\subsubsection{DEM}
Additional elevation information improved the performance by the comparisons of trial 3/4 versus trial 1, and trial 10/11 versus trial 8. Elevation information alone, however, only worked for the detection and segmentation of large rocks, which may be caused by the comparable-size bushes in the site, e.g. Fig.~\ref{fig:rgb_dem_ndvi}. Relative elevation (DEM1) outperformed colormap elevation (DEM3) in trial 3 versus trial 4, trial 10 versus trial 11, and trial 13 versus trial 14. Though from trial 6 versus trial 7 these two elevation representations had comparable improvements, relative elevation showed slight advances in key metrics.

\subsubsection{Multispectral orthomosaics}
RE and NIR can reveal information that RGB alone cannot obtain. For example, normalized difference vegetation index (NDVI) has largely been used for vegetation detection. An example of NDVI tiles is shown in Fig.~\ref{fig:rgb_dem_ndvi}(D). While NDVI was not directly used in our experiment, we included original RE and NIR in the input and let the neural networks mine the multispectral orthomosaics themselves. RE+NIR assisted neural network performance when added to RGB, RGB+DEM3, and RGB+DEM1 in both transfer learning and non-transfer learning. 

\subsubsection{Transfer learning}
The trials with transfer learning generally outperformed the ones without transfer learning with the exception of some acceptable noises in $AP_2$. Considering key metrics, transfer learning did demonstrate advances in the inference performances. 

\subsubsection{DEM and Multispectral orthomosaics} RGB+RE+NIR+DEM3/RGB+RE+NIR+DEM1 yielded better results than other trials. While in transfer learning RGB+RE+NIR+DEM1 was slightly better at key metrics, it surpassed other trials in non-transfer learning.

\section{Contribution and Future Work}
In this paper, we presented a pipeline for geomorphological analysis using structure from motion and deep learning on close-range aerial imagery. Our UAS-SfM-DL pipeline was used to assess the effectiveness of multispectral data and elevation representations in neural networks and to estimate the distribution of rock traits (size, roundness, and orientation) on a fault scarp in the Volcanic Tablelands, California. Although presented in the context of fault zone geology, we foresee our pipeline being extended to a variety of geomorphological analysis tasks in other domains such as crop property estimation in precision agriculture and debris field analysis after natural disasters.

From our first, proof-of-concept experiment, we conclude that relative elevation improved the neural network performance in both transfer learning and non-transfer learning. With transfer learning, we have shown both elevation representations assisted neural network performance from RGB data, and the relative elevation resulted in the best improvement. We also obtained rock-trait histograms along and across the fault scarp. The distributions of rock size are asymmetrical throughout the fault, with larger rocks in the north and smaller rocks in the south. Additionally, the larger particles are typically in the mid scarp position. Such information can guide future scientific inquiries about strain across the fault as well as geomorphic modifications of fault scarps to better understand earthquake recurrence in a tectonically active region. As far as we know, this is the first time that UAVs and machine learning are used to measure rock traits on fault scarps.

In the second experiment, we focused on examining the effectiveness of different input combinations of multispectral orthomosaics and two elevation representations. From the inference results, additional spectral data and elevation information improved the performance of the neural networks. Even though the first convolutional layer of Mask R-CNN needed to be retrained, transfer learning showed general advances over non-transfer learning in all combination settings.  

Field inspection of the fault scarp indicated that the grain size of the rocks included those smaller than the mode of $~$0.2 meters (Fig.~\ref{fig:rock-size-hist}(C)). The rollover in grain size to the smaller side is therefore likely due to a sensitivity issue from both image resolution (2-cm pixels for orthomosaics) and neural network architectures. Addressing the finer size tail on the rock size distribution is an important topic for future research.

Rock sizes are approximated using 2D areas from orthomosaics. We will look at rock size estimation from 2.5D (elevation) and 3D (point cloud) approaches. In this work, we simply stack all different data as input and implicitly rely on neural networks to learn useful features from input channels for instance segmentation. We will investigate attention mechanism to actively weight interesting input channels, which will benefit deep neural network learning in multispectral and hyperspectral data.

\section*{ACKNOWLEDGEMENTS}
This work was supported in part by Southern California Earthquake Center (SCEC) award 19179, National Science Foundation award CNS-1521617, and National Aeronautics and Space Administration STTR award 19-1-T4.01-2855.  Thank you to Duane DeVecchio for discussion on the geomorphic concepts presented here.


{\small
\printbibliography}

\end{document}